\documentclass{ieeeaccess}
\usepackage{cite}
\usepackage{amsmath,amssymb,amsfonts}
\usepackage{algorithmic}
\usepackage{graphicx}
\usepackage{textcomp}

\usepackage{bm}
\makeatletter
\AtBeginDocument{\DeclareMathVersion{bold}
\SetSymbolFont{operators}{bold}{T1}{times}{b}{n}
\SetSymbolFont{NewLetters}{bold}{T1}{times}{b}{it}
\SetMathAlphabet{\mathrm}{bold}{T1}{times}{b}{n}
\SetMathAlphabet{\mathit}{bold}{T1}{times}{b}{it}
\SetMathAlphabet{\mathbf}{bold}{T1}{times}{b}{n}
\SetMathAlphabet{\mathtt}{bold}{OT1}{pcr}{b}{n}
\SetSymbolFont{symbols}{bold}{OMS}{cmsy}{b}{n}
\renewcommand\boldmath{\@nomath\boldmath\mathversion{bold}}}
\makeatother

\def\BibTeX{{\rm B\kern-.05em{\sc i\kern-.025em b}\kern-.08em
    T\kern-.1667em\lower.7ex\hbox{E}\kern-.125emX}}

\usepackage{url}
\usepackage{algorithm}

\usepackage{booktabs}
\usepackage{multirow}
\usepackage{graphicx}
\usepackage{caption}
\usepackage{subcaption}
\usepackage{amsmath}
\usepackage{cleveref}
\usepackage{amssymb}
\usepackage{pifont}

\newcommand{\cmark}{\ding{51}}%
\newcommand{\xmark}{\ding{55}}%

\def\modelOne{CAB1}
\def\modelTwo{CAB2}

\begin{document}
\history{Date of publication xxxx 00, 0000, date of current version xxxx 00, 0000.}
\doi{}

\title{Recognizing and Reconstructing a Multi-Unit Floor Plan}

\author{\uppercase{Lukas~Kratochvila}\authorrefmark{1},
\uppercase{Gijs~de~Jong}\authorrefmark{2}, Monique~Arkesteijn\authorrefmark{3}, Tomas~Zemcik\authorrefmark{1}, Simon~Bilik\authorrefmark{4}, Karel~Horak\authorrefmark{1}, and~Jan~S.~Rellermeyer\authorrefmark{5}}

\address[1]{Department of Control and Instrumentation, Brno University of Technology, Brno, Czech Republic. E-mail: kratochvila@vut.cz.}
\address[2]{Department of Software Technology, Faculty of Electrical Engineering Mathematics and Computer Science, TU Delft, Delft, Netherlands.}
\address[3]{Department of Management in the Built Environment, Faculty of Architecture and the Built Environment, TU Delft, Delft, Netherlands.}
\address[4]{Institute for Research and Applications of Fuzzy Modeling, University of Ostrava, Ostrava, Czech Republic and with Department of Informatics, Mendel University in Brno, Brno, Czech Republic.}
\address[5]{Department of Software Technology, Faculty of Electrical Engineering Mathematics and Computer Science, TU Delft, Delft, Netherlands and with Dependable and Scalable Software Systems, Institute of Systems Engineering, Faculty of Electrical Engineering and Computer Science, Leibniz University Hannover, Hannover, Germany.}

\tfootnote{The completion of this paper was made possible by the grant No. FEKT-S-26-8988 - ”Advanced Methods in Cybernetics, Robotics, and Artificial Intelligence” financially supported by the Internal science fund of Brno University of Technology.}

\markboth
{Author \headeretal: Preparation of Papers for IEEE TRANSACTIONS and JOURNALS}
{Author \headeretal: Preparation of Papers for IEEE TRANSACTIONS and JOURNALS}

\corresp{Corresponding author: Lukas Kratochvila (e-mail: kratochvila@vut.cz.)}

\begin{abstract}
Digital twins have a major potential to form a significant part of urban management in emergency planning, as they allow more efficient designing of the escape routes, better orientation in exceptional situations, and faster rescue intervention. Nevertheless, creating the twins still remains a largely manual effort, due to a lack of 3D-representations, which are available only in limited amounts for some new buildings. Thus, in this paper we aim to synthesize 3D information from commonly available 2D architectural floor plans. We propose two novel pixel-wise segmentation methods based on the MDA-Unet and MACU-Net architectures with improved skip connections, an attention mechanism, and a training objective together with a reconstruction part of the pipeline, which vectorizes the segmented plans to create a 3D model. The proposed methods are compared with two other state-of-the-art techniques and several benchmark datasets. On the commonly used CubiCasa benchmark dataset, our methods have achieved the highest F1 score of 0.86, outperforming the other approaches tested. We have also made our code publicly available to support research in the field.
\end{abstract}


\titlepgskip=-21pt

\maketitle

\section{Introduction}
\label{sec:introduction}
\PARstart{U}{rban}  areas experience a steady growth, and over two-thirds of the population worldwide will be considered urbanized by the year 2050~\cite{desa2018world}. Heavy and complex traffic flows and multiple buildings of various ages and structural arrangements make emergency planning against events such as a fire or flood very challenging. Even though the emergency operators usually have access to the building schemes and floor plan drawings, the individual embodiments of these documents may vary markedly, meaning that, for instance, the paper sheet rendering differs textually and graphically from that of the electronic version (such as a PDF file). Quick understanding of the 2D drawings during a rescue task might then be challenging due to not only these differences but also a high number of the records and their lower clarity compared to the 3D space representation.

A promising way to change this situation lies in using the digital twin (DT) of the relevant city, as demonstrated on the examples of Cambridge campus, ~\cite{qiuchen2019developing} and an office building facade element,~\cite{khajavi2019digital}; the approach is also discussed in an extensive overview centered on the individual aspects and applications,~\cite{jones2020characterising}. The DT models could become an important part of urban management, improving the response to emergencies and making safety planning more direct and efficient. The majority of existing DT systems, however, were developed only through simulation or for recently built structures with an existing 3D model.

The research presented in this paper has been conceived to close the above-outlined gap by formulating a novel procedure to generate automatically a 3D representation of buildings from conventional 2D-floor plans, which are usually available from the competent municipal authorities. As the option proposed herein only estimates the height of the floors, it could be termed a 2.5D model by some authors. The 3D models are seamlessly integrable into the digital twins to model entire cities in a scalable and economical manner, unlike using a point cloud as shown, for example in~\cite{BUT183869}, which is computationally very demanding and with noisier results. In generating such a 3D model from the 2D plans, an essential issue rests in reliable recognition of critical structural elements in floor plans, including walls, windows, stairs, and railings. While numerous approaches are in use presently, a substantial number of these assume the 2D representation to already come in the form of vectors~\cite{Fan_2021_ICCV}, a case not typical of older buildings. Other approaches do not rely on this assumption but aim merely at relatively small-sized plans of apartments and smaller units, not covering entire multi-unit floors or buildings. The central goals of the research can be summarized as follows:

\begin{itemize}
    \item developing a novel recognition and reconstruction pipeline that involves semantic-wise pixel segmentation, recognition of structural elements, and 3D reconstruction;
    \item evaluating the proposed segmentation method with other state-of-the-art techniques;
    \item testing the method on various datasets consisting of multi-unit floor plans and various drawing standards;
    \item publishing the framework to support further investigation in the domain.\footnote{Our code: \url{https://github.com/LukasKratochvila/multi-unit-floorplan}}.
    
\end{itemize}

\section{Related Work}
\label{sec:background}
Automated recognition of architectural plans and drawings embodies a challenging problem which may benefit multiple fields and subdomains, such as creating virtual twins of buildings~\cite{qiuchen2019developing}, optimal evacuation path planning~\cite{smartcities6040084}, and firefighting. As applicable architectural plans are often unavailable electronically or do not exist at all, digitizing older drawings or the common floor and evacuation schemes might be a necessary precondition for this task.

The first attempts to facilitate automated recognition of architectural plans and drawings were made more than 30 years ago, as presented in paper~\cite{201647}. Other procedures, utilizing classic computer vision techniques such as segmentation, recognition, and classification, appeared at the beginning of the new millenium; these options included, for instance, innovative systems ~\cite{dosch2000complete}, binary image operations~\cite{or2005highly,ahmed2011improved}, the Hough transformation ~\cite{mace2010system}, and a combined approach comprising structural and semantic analysis complemented with OCR-retrieved information~\cite{ahmed2012automatic}. Although other segmentation methods have been designed ~\cite{Bostik_Valach_Horak_Klecka_2019}, we focus especially on advances in convolutional neural networks (CNNs) after the introduction of AlexNet in 2014~\cite{NIPS2012_c399862d}: As deep learning-based techniques have found use in executing the tasks characterized herein, proving to be very efficient in semantic segmentation, we decided to opt for deep learning.

A CubiCasa5K-related paper~\cite{kannala2019cubicasa5k} introduces a new, large dataset of 5,000 floor plans with more than 80 annotated classes; by extension, an original multi-task model is outlined. Exploiting the previously published segmentation network from~\cite{liu2017raster}, the author adds a trainable module to adjust the weights in the multi-task loss calculation.

Paper~\cite{lv2021residential} exposes a complex framework combining wall segmentation based on DeepLabV3+~\cite{chen2018encoder} and opening detection with a YOLOv4~\cite{bochkovskiy2020yolov4} object detector. The results of the research comprise a novel scale calculation relying on floor plan annotations to determine the size of a 3D representation. The whole method was tested on an in-house compiled dataset. Compared to the solution in~\cite{zeng2019deep}, the approach increases the accuracy of both the wall and the room segmentation types.

The authors of~\cite{Vidanapathirana2021Plan2Scene} discuss a tool named plan2scene to produce textured 3D models of residential interiors using vector floor plans and a sparse set of photographs for the relevant texture. The focus is on texture generation and propagation, namely, extracting textures from observed data via an encoder-decoded architecture (as opposed to the previously employed approach, which requires precise calibration of the images to the model) and producing more generic textures to be propagated onto unobserved parts of the model by applying a GNN (Graph Neural Network)-based model. The research project outlines metrics to judge the performance of a textured model generation, claiming capabilities that surpass those of comparable models.

In~\cite{electronics10222729}, the aim lies in converting a 2D raster drawing into 3D data through a deep neural network technique on a limited dataset of 30 floor plans, utilizing the 3DPlanNet. To achieve this, they authors use a complex pipeline consisting of pattern recognition with a moving kernel to enable wall detection; object detection in corners; wall junction; and door, window, and room type recognition using the TensorFlow detection API followed by the generation of nodes, edges, and detected objects such as doors and windows. The approach was tested on a KOVI dataset, which, regrettably, is not described in greater detail.

The research set out in~\cite{9893304} concentrates on an automatic method for segmenting raster-wise floor plans by means of learning-based hierarchical segmentation and assembling the obtained geometrical primitives into a planar structure. A 3D model generation from the floor plan is included in the proposed framework. A VGG~\cite{simonyan2014very} classifier to handle the objects and boundaries is followed by a custom semantic segmentation network with a loss function, respecting the boundaries and type of the room. The obtained geometrical primitives are identified and fused with the relation information into a planar graph using mixed integer programming. The proposed approach was tested on R2V~\cite{liu2017raster} and R3D~\cite{liu2015rent3d} datasets.

\section{Proposed Methods}
\label{sec:method}

\begin{figure*}[t!]
    \centering
    \includegraphics[width=0.6\textwidth]{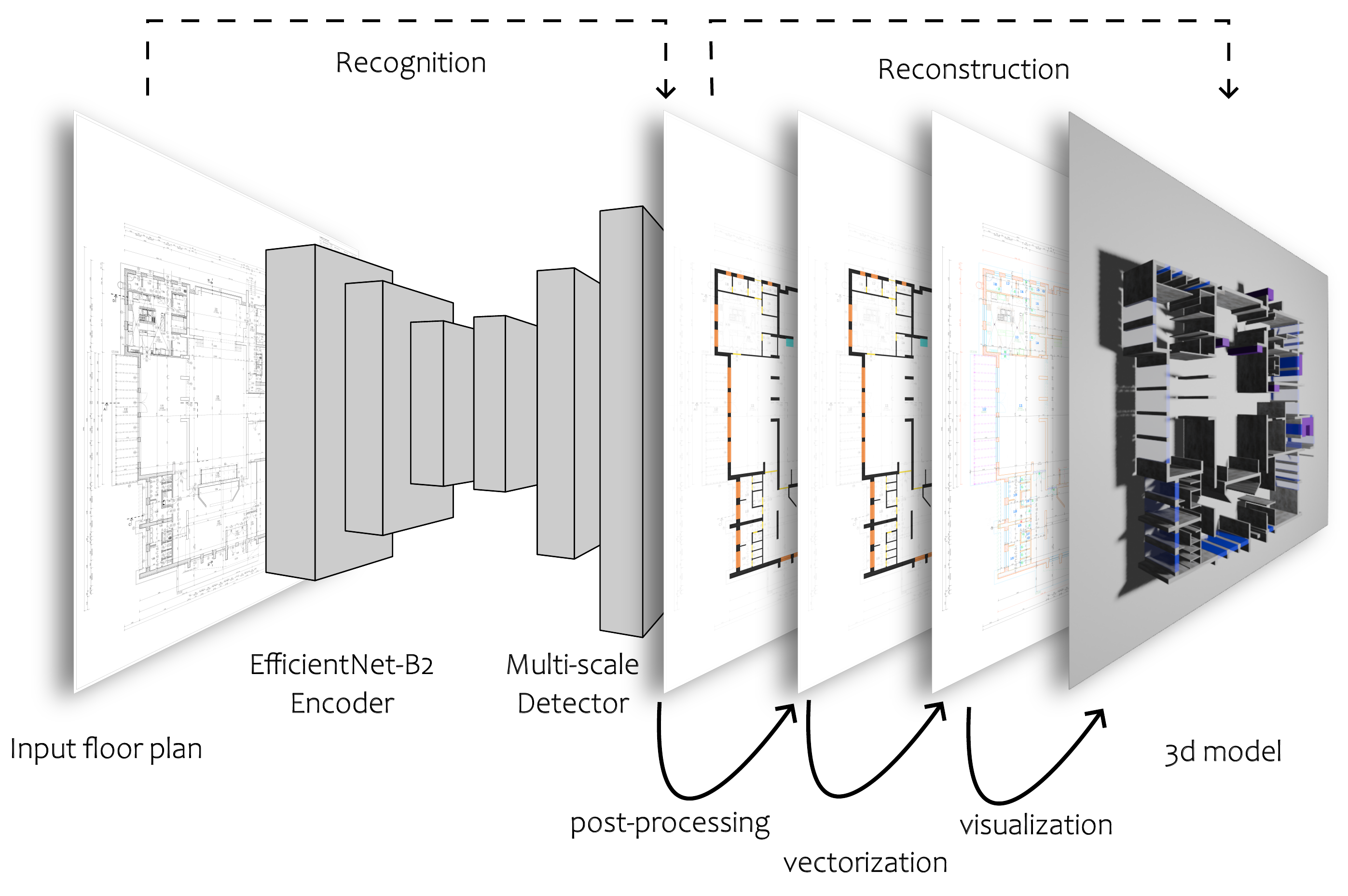}
    \caption{The recognition part, exposed on the left-hand side, produces a segmentation mask of the floor plan by using a custom convolutional neural network (CNN). The segmentation mask is subsequently refined through the reconstruction step, shown on the right-hand side, which applies post-processing, vectorization, and Blender-based visualization.}
    \label{fig:overview}
\end{figure*}

The structure of the designed floor plan processing pipeline is shown in~\Cref{fig:overview}. The unit can be  separated into a recognition part (on the left), which performs a pixel-wise semantic segmentation of the given floor plan, and a reconstruction sector (on the right), which further processes the segmented elements into a final 3D model.

\subsection{Recognition}

\begin{figure*}[t!]
    \centering
    
    \subfloat[A model \modelOne{} with regular multi-scale feature skip connections.\label{fig:model_regular}]{%
        \includegraphics[width=0.4\textwidth]{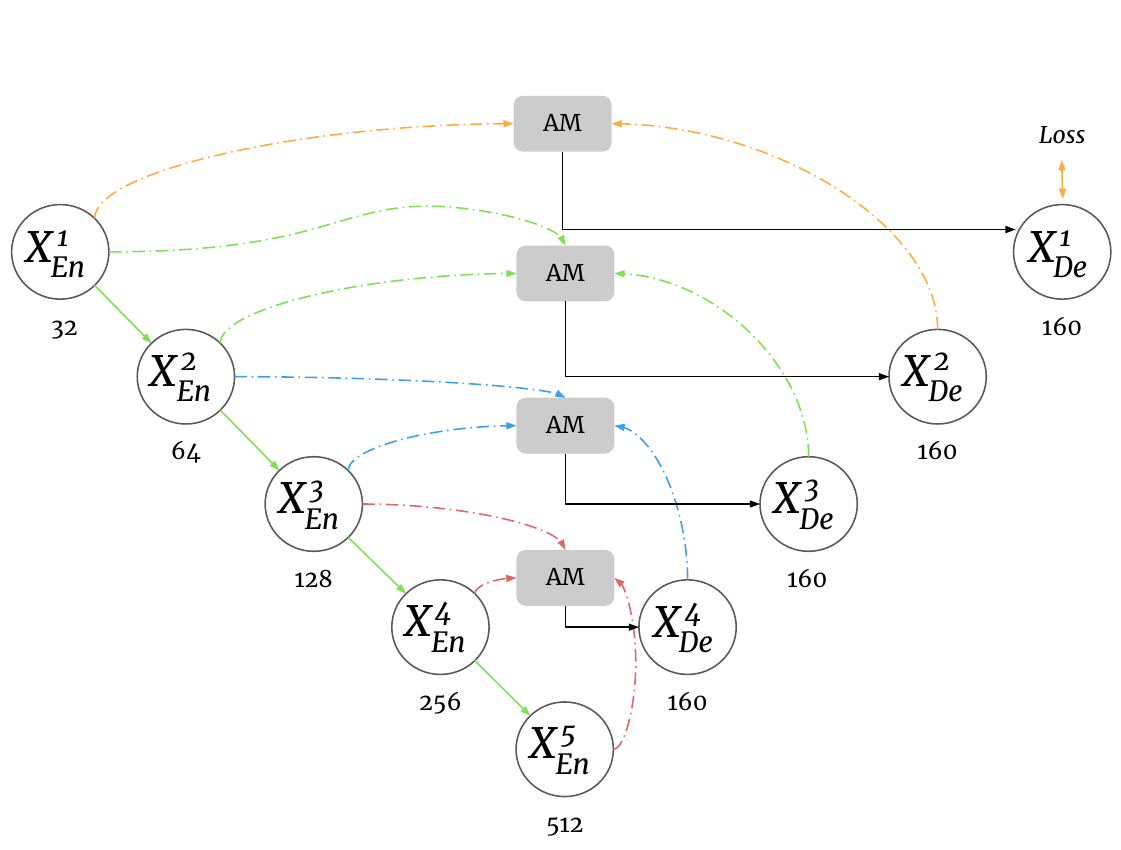}%
    }
    \hfill
    \subfloat[A model \modelTwo{} with fully connected multi-scale feature skip connections.\label{fig:model_fully}]{%
        \includegraphics[width=0.4\textwidth]{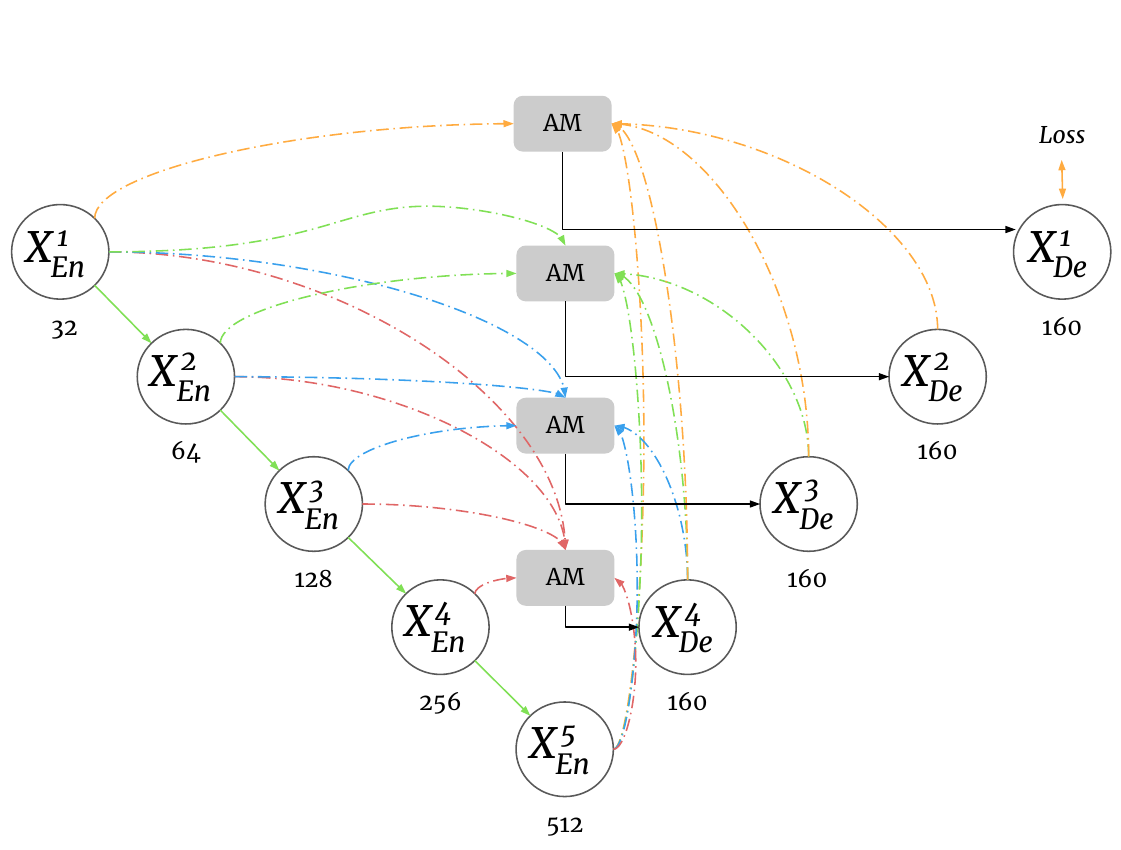}%
    }
    
    \vspace{10pt} 
    
    \subfloat[\label{fig:model_legend}]{%
        \includegraphics[width=0.4\textwidth]{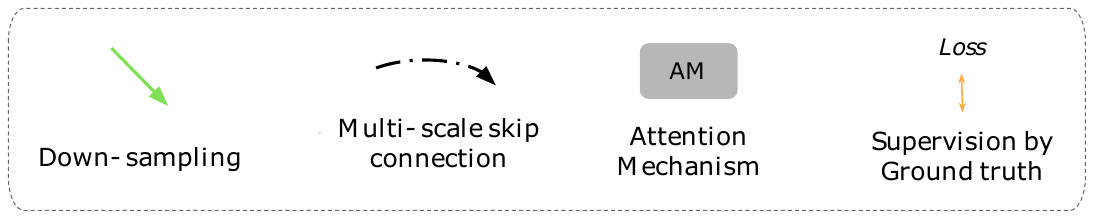}%
    }
    
    \caption{Two model architectures: (a) regular multi-scale feature skip connections, and (b) fully connected multi-scale feature skip connections. Here, $X^i_{En}$ means an encoder and $X^i_{De}$ a decoder block.}
    \label{fig:models}
\end{figure*}

The recognition part exploits a custom CNN based on an MDA-UNet~\cite{amer2022mda}, MACU-Net~\cite{li2020macu}, and  U-Net3+~\cite{qin2022improved}; by extension, it presents an asymmetric convolution block, an attention mechanism, a multi-scale feature skip connections, and a multi-task training objective. The training objective is designed to capture a pixel, a patch, and multi-scale losses. \Cref{fig:models} depicts the two model architectures and their structural properties. In addition, we included different input image sizes to the training process to create a multi-scale detector that can process variously sized input images. 

    
    

\subsubsection{Asymmetric Convolution Block}
\begin{figure*}[!t]
    \centering
    \includegraphics[width=0.8\textwidth]{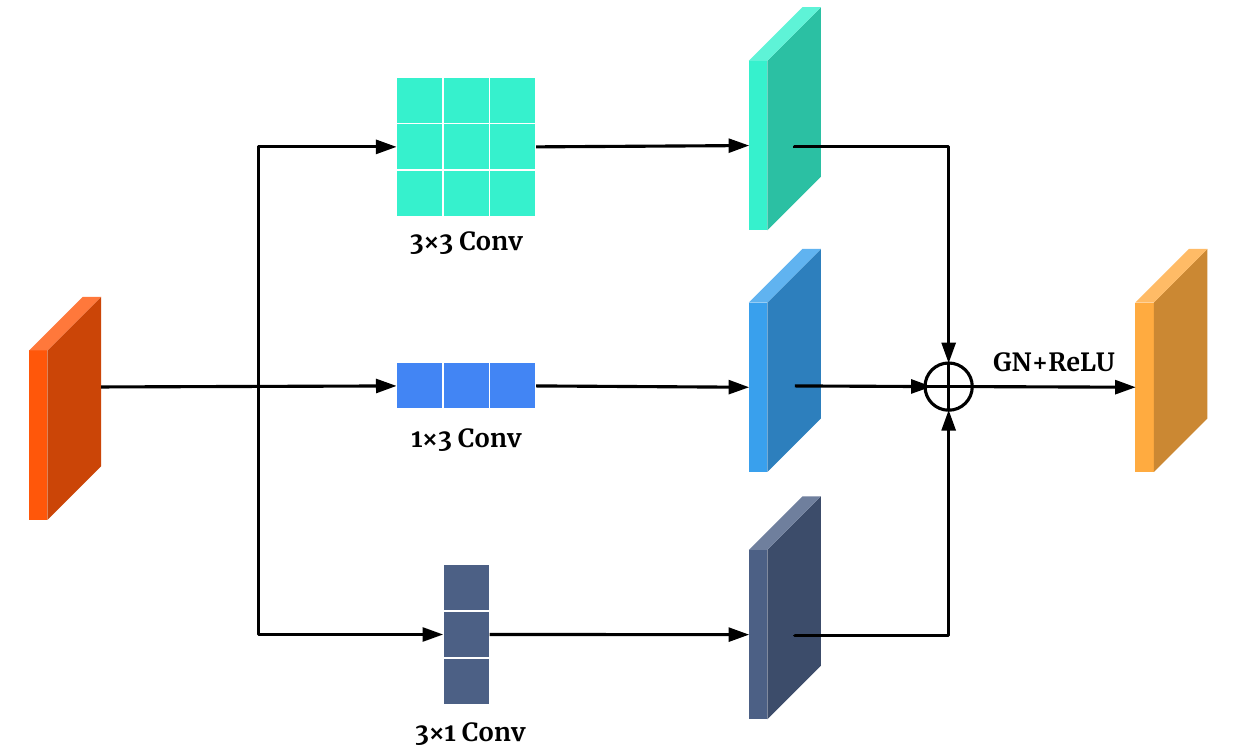}
    \caption{An asymmetric convolution (AC) block enhancing the skeleton features; $\bigoplus$ represents element-wise addition, and GN stands for group normalization.}
    \label{fig:ac}
\end{figure*}

To better capture the learned feature maps, different types of convolution blocks are applicable. In addition
to the normalization and activation functions paired with the convolution layers, the design and use of
kernels play an important role in improving the performance of a CNN \cite{ding2019acnet}. We propose an Asymmetric Convolution (AC) block, as shown in~\Cref{fig:ac}, to replace the standard square convolutions. Asymmetric convolution blocks have been shown to enhance the skeleton features by replacing the standard k × k (k denotes the kernel size) kernels with an aggregation of the k × k kernel and two additional horizontal 1 × k and vertical k × 1 kernels \cite{ding2019acnet}. Directional kernels have also been evaluated on floor plans; however, they do not deliver clear performance improvements compared to horizontal and vertical kernels \cite{zeng2019deep,zhang2020direction}.

\subsubsection{Attention Mechanism}

Each of the proposed models utilizes an attention mechanism that combines a channel attention module with a spatial attention module to highlight the feature maps separately, based on their spatial or channel importance, similarly to the design proposed by~\cite{woo2018cbam}. The attention module infers a channel attention map $M_c$ and a spatial attention map $M_s$ from an intermediate feature $F$, which is subsequently weighed through multiplication formalized by \Cref{eq:am}. The final refined feature is computed by performing an asymmetric convolution on the concatenated channel and spatial refined feature.

\begin{equation}
    \label{eq:am}
    \begin{split}
       AM(F)&=AC \biggl( \biggl[ F_c \otimes M_c(F), F_s \otimes M_s(F) \biggr] \biggr),\\
       F_c &= C^{1\times1}(F),\\
       F_s &= C^{1\times1}(F),
    \end{split}
\end{equation}

where $F_c$ and $F_s$ are the channel compressed representations of the intermediate feature map $F$ using two $1\times1$ convolutions. Note that the spatial map $M_s$ is derived from the full-sized feature $F$. The concatenation is denoted by square brackets, and $\otimes$ denotes the element-wise multiplication.

Concatenating average and max-pooled feature maps has been shown as an effective step in reinforcing spatial features~\cite{amer2022mda}. Different dilation rates have been used to capture the features from different-sized receptive fields, reinforcing small and large spatial features. The spatially enhanced features are aggregated through element-wise addition and normalized by the sigmoid function to produce the final spatial map. Spatial and channel attention maps are discussed in more detail in the supplementary materials.

\subsubsection{Multi-Scale Feature Skip Connections}

\begin{figure*}[!t]
    \centering
    \includegraphics[width=0.8\textwidth]{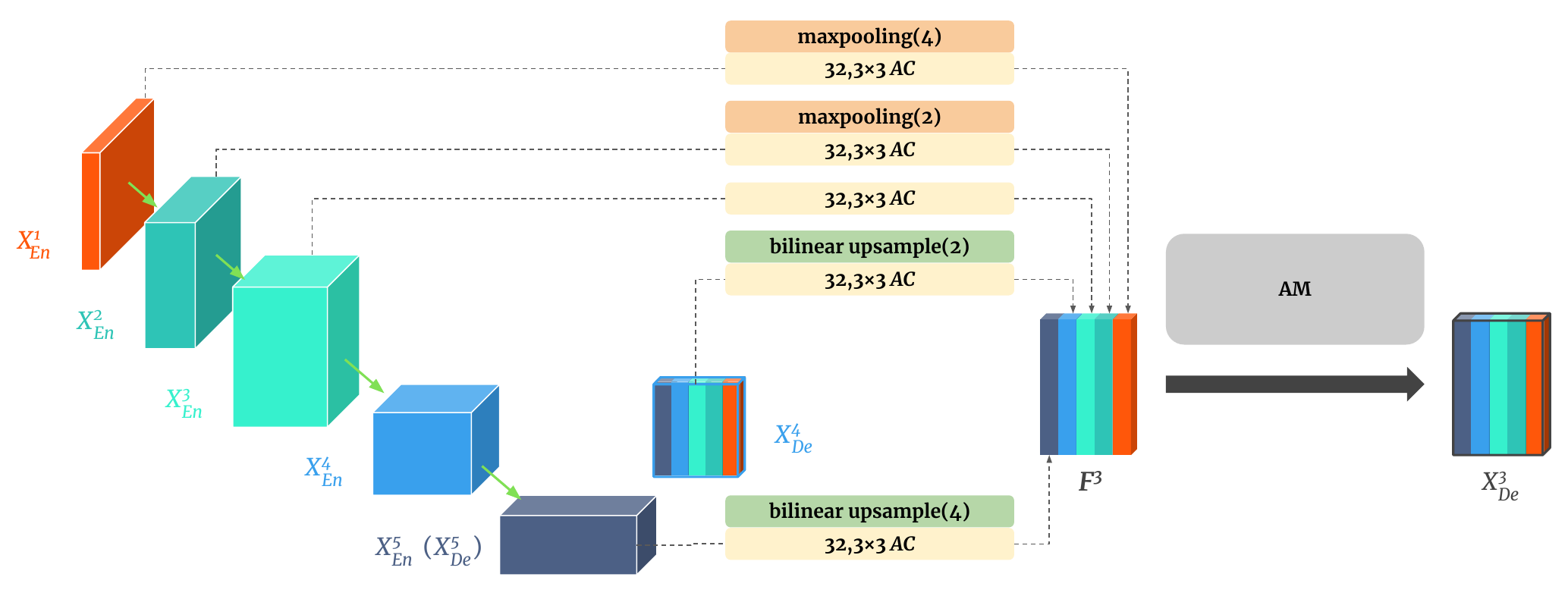}
    \caption{An example of an intermediate feature map $F^3$ of the third decoder layer $X^3_{De}$. The abbreviations AC and AM indicate an asymmetric convolution block and an attention mechanism;
    $X^i_{En}$ denotes blocks of convolutions.}
    \label{fig:am}
\end{figure*}

As the multi-scale feature skip connections have recently provided promising results, such as mitigating the disadvantages of the full-scale skip connections o the U-Net3+, our architectures incorporate them. Below, \Cref{fig:am} shows the construction of $X^3_{De}$. The intermediate feature map $F$ of the model architectures is structured similarly to that of the U-Net3+. For each up-sampling level $i<N$ and respective decoder $X^i_{De}$, the feature map $F^i$ to be refined comprises a mix of differently scaled feature maps from the encoder and the decoder layers. 

As shown in~\Cref{fig:models}, the first architecture \modelOne{} contains regularly connected multi-scale feature skip connections, where the individual encoder layers are combined with the previous (shallower) layer of the encoder. In the case of a decoder, the layers are combined with encoder layers of the same or previous depth and the previous (deeper) layer of the decoder. 

The second architecture, \modelTwo{}, uses the same encoder as the model \modelOne{}; however, the decoder layers are combined with all previous (shallower) encoder layers and all previous (deeper) decoder ones.

\subsubsection{Training Objective}

Further, we propose a multi-task training objective, with each task designed for a specific purpose, namely, handling class imbalance, utilizing semantic relations in labels, and exploiting multi-scale predictions. We employ an asymmetric unified focal loss to handle class imbalance by suppressing the background loss and enhancing the foreground loss as described in~\cite{yeung2022unified}.

The unified focal loss applies a form of unary supervision, which lacks the spacial discrimination to exploit the semantic structure in the labels. To capture this structure, we use the affinity field loss defined in~\cite{ke2018adaptive}. 

The third loss, formalized in~\Cref{eq:heatmap_loss}, represents a heatmap regression task similar to that presented in~\cite{lv2021residential}. This loss is of a distinct character, as the heatmaps are indirectly derived from the segmentation masks instead of being predicted by the model. Regarding the heatmaps, the heatmap regression loss $\mathcal{L}_{\text{MHR}}$ is a simple average of the squared euclidean distance between the prediction map $\hat{y}$ and the heatmap $H_{c}$ over all the openings O and pixels i in the floor plan. $\mathcal{O}_c$ is the set of endpoints of opening c, $y$ is ground truth image, and $\beta$ controls the spread of the peak of the values in the heatmap. More details about the heatmaps are provided in the supplementary materials.

\begin{equation}
    \begin{split}
    \mathcal{L}_{\text{MHR}} &= \frac{1}{|O|}\sum_c^O\frac{1}{|\hat{y}|}\sum_i||\hat{y}_c(i)-H_{c}(i)||_2^2
    \\
    \text{where:}\hspace{25pt}&\\
    H_c(i) &= \frac{1}{|B|}\sum_\beta^B\bigg(\max_{j\in \mathcal{O}_c}\exp(-\frac{||y_i(c)-y_j(c)||_2^2}{\beta^2})\bigg)
    \end{split}
    \label{eq:heatmap_loss}
\end{equation}

The final multi-task loss is the weighted sum of all previously defined loss functions, for which the weights are learned through training by using the approach proposed in~\cite{liebel2018auxiliary}.

\subsection{Reconstruction}
\label{sec:reconstruction}

The reconstruction part of the pipeline refines the segmentation masks from the recognition sector by applying three steps: post-processing, vectorization, and visualization. The post-processing method continuously refines the connected components in the segmentation mask until an uncertainty threshold is reached. Such refined components of the segmentation mask are transformed into polygons by deriving the endpoints of the minimum area rotated rectangle of each component. The polygons are then refined by merging the vertices that are sufficiently close or collinear. The resulting set of polygons is used to visualize the segmentation mask with the Blender.

\begin{algorithm}
    \caption{Polygon Approximation}\label{alg:poly}
    \begin{algorithmic}
        \REQUIRE Joint mask $\hat{y}$, uncertainty threshold $\epsilon_u$
        \STATE Find the initial connected components $C$ in $\hat{y}$, using~\cite{bolelli2019spaghetti}
        
        \WHILE{Any $C$ left}
            \STATE Find a rotated rectangle $bb$ with the minimum area enclosing the component~\cite{toussaint1983solving}
            \STATE Calculate the uncertainty $u_{bb}$ of $bb$ according to \Cref{eq:pp_uncertainty}
            
            \IF{$u_{bb}$ is smaller than $\epsilon_u$}
                \STATE Divide the $bb$ into two smaller rectangles as shown in \Cref{fig:bb_divide}
                \STATE Add a new $C$ of each new rectangle
            \ELSE
                \STATE Add the approximate polygon $P_i$ to the list of polygons $P$
            \ENDIF
        \ENDWHILE
    \end{algorithmic}
\end{algorithm}

\subsubsection{Approximate Polygons}

The first step to derive the approximate polygons from the segmentation mask rests in creating a joint mask consisting of the union of all the semantic classes. Such a joint mask is proceeded by \cref{alg:poly}:

The $\epsilon_u$ is a constant uncertainty threshold set to $0.5$ determined empirically. The value was found to be high enough to produce accurate approximations and low enough to generalize excessively noisy predictions. The uncertainty $u_{bb}$ is defined below:

\begin{equation}
    u_{bb} = \biggl( \frac{1}{|bb|} \sum_{i\in bb} \hat{y}(i) \biggl) 
    \label{eq:pp_uncertainty}
\end{equation}


\begin{figure*}[!tb]
    \centering
    \subfloat[Initial bounding rectangle.]{%
        \label{fig:bb_full}
        \includegraphics[width=0.26\textwidth]{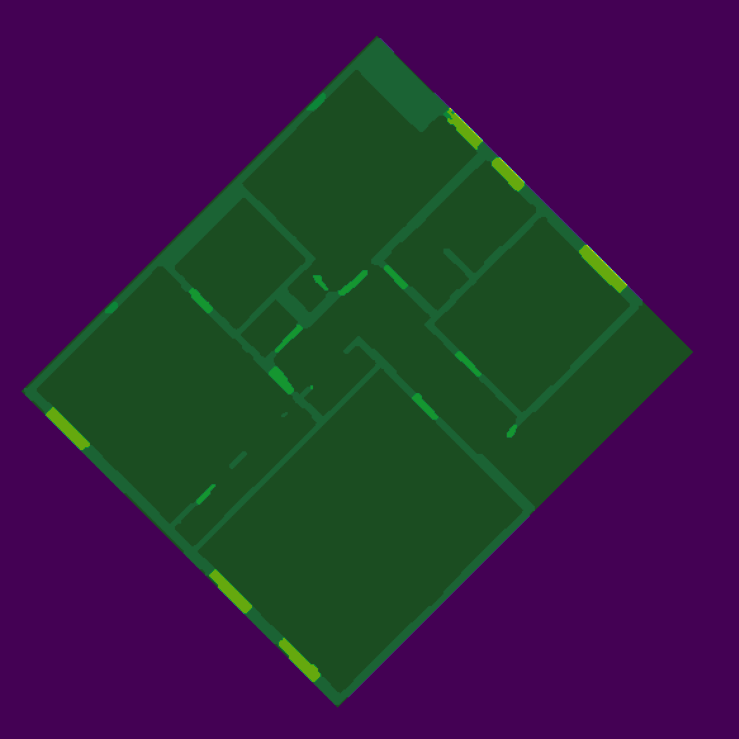}
    }
    \hfill
    \subfloat[First new bounding box of the first half.]{%
        \label{fig:bb_half1}%
        \includegraphics[width=0.26\textwidth]{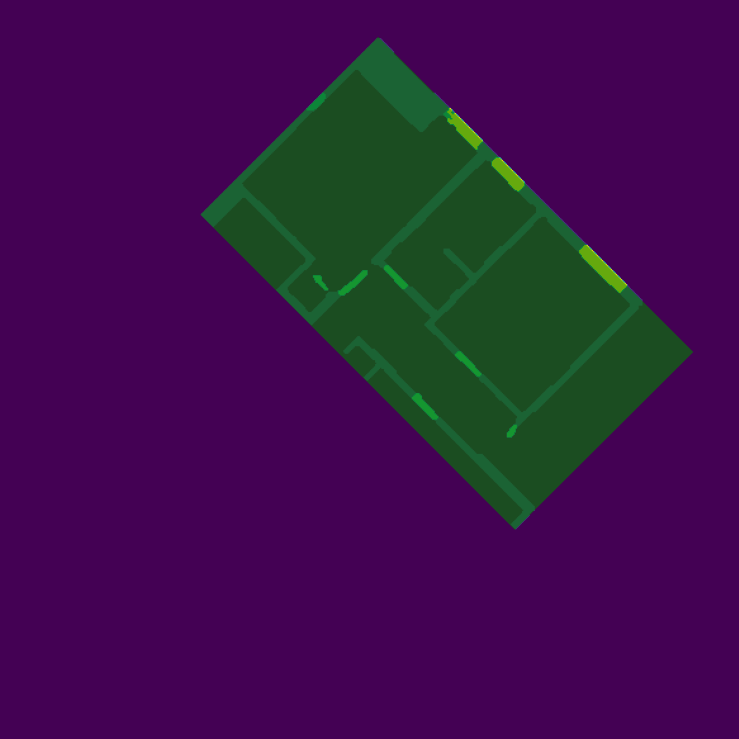}%
    }
    \hfill
    \subfloat[Second new bounding box of the second half.]{%
        \label{fig:bb_half2}%
        \includegraphics[width=0.26\textwidth]{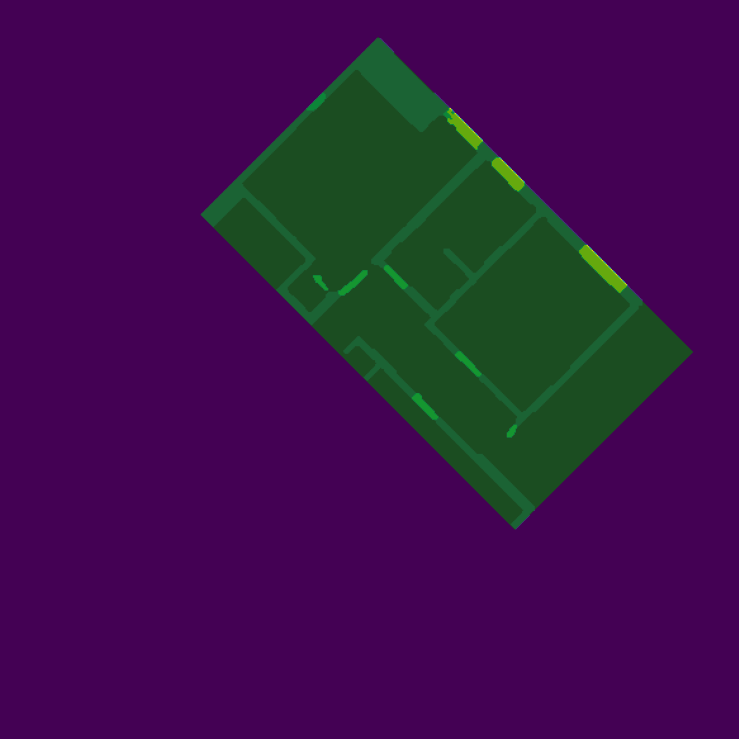}%
    }
    \caption{An example of the bounding box, and smaller bounding boxes generated to derive new components. The smaller bounding boxes divide the largest side of the initial bounding box in two.}
    \label{fig:bb_divide}
\end{figure*}

\subsubsection{Refined Polygons}

The refined polygons are obtained by reducing the approximate ones in two steps according to ~\cite{lv2021residential}. The first step is performed to merge all vertices whose Euclidean distance is smaller than or equal to the pre-defined distance threshold $\epsilon_d$. The second step then refines the polygons by removing vertices approximately collinear to two other vertices. A vertex is considered collinear to two other vertices if the angle between them is greater than or equal to the threshold $\epsilon_a$. The distance and angle thresholds $\epsilon_d$ and $\epsilon_a$ have been found empirically and set to $\epsilon_d = 4$ and $\epsilon_a = \cos(14^{\circ})$. Refining the polygons is described in \cref{alg:ref}.

\begin{algorithm}
    \caption{Polygon Refining}\label{alg:ref}
    \begin{algorithmic}
        \REQUIRE Set of polygons $P$, thresholds $\epsilon_d$, $\epsilon_a$
        \STATE Construct the initial KD-tree of vertices $V$ from all $P$ and create an empty set $V_c$ of vertices to be merged
        \FOR{All $v_i\in V$}
                \STATE To a set $V_c$ add $v_j$ if $v_i\in P_i$, $v_j\in P_j$ and $P_i \neq P_j$ and $||v_i-v_j||^2_2 \leq \epsilon_d$
        \ENDFOR
        \WHILE{Any $v_j\in V_c$}
            \STATE Replace $v_i\in V_j$ and $v_j \in V_c$ with $\frac{v_i + v_j}{2}$ in the respective polygons $P_i$ and $P_j$
            \STATE Construct a new KD-tree of vertices $V$ from all $P$
            \FOR{All $v_i\in V$}
                \STATE Compute the set $V_c$, $v_i\in P_i$, $v_j\in P_j$ and $P_i \neq P_j$ and $||v_i-v_j||^2_2 \leq \epsilon_d$
            \ENDFOR
        \ENDWHILE    
        \WHILE{Any $|\cos( \overrightarrow{v_iv_j}, \overrightarrow{v_jv_k} )| \geq \epsilon_a$ for $v_i$, $v_j$, $v_k \in P_i$}
            \STATE Remove the middle vertex $v_j$
        \ENDWHILE
    \end{algorithmic}
\end{algorithm}

\section{Experimental Description}
\label{sec:experiments}

In this paper, we first compare the proposed segmentation models \modelOne{} and \modelTwo{} of our pipeline with the DFPR~\cite{zeng2019deep} and Cubicasa5k~\cite{kannala2019cubicasa5k} SOTA methods on the Cubicasa dataset. Further, the \modelOne{} and \modelTwo{} models are evaluated on the Cubicasa, R3D, and CVC-FP datasets. We decided to employ the DFPR and Cubicasa5k methods, as they are well described and, with the implementations, publicly available. The Cubicasa, R3D, and CVC-FP datasets are ones with publicly available data, have different classes, and are evaluated along a multiple methods.

\subsection{Model Modifications}

To ensure a fair comparison between the evaluated models, we modified the DFPR and Cubicasa5k by removing the fixed input size requirements and room predictions. The room predictions were removed because they would otherwise only reduce the accuracy of the boundary and opening classes. Concerning the Cubicasa5k model, the balanced entropy loss function from the DFPR model was adopted, as the original categorical cross-entropy loss failed to converge. The Cubicasa5k model is an improved version of~\cite{liu2017raster}; this model is therefore indirectly evaluated as well. The edited models are noted as the DFPR$_\beta$ and the CubiCasa5k$_\beta$. A more detailed description of these modifications is provided in the supplementary materials.

\subsection{Dataset Description}

As mentioned above, several datasets associated with floor plan analysis are available. Below, \Cref{t:FPDatasets} compiles some of the existing datasets that were deemed suitable for the discussed task. The suitability is based on the nature of the floor plan data plus the format and the extent of the data annotation.

\begin{table*}[ht!]
	\begin{center}
        \begin{minipage}{\textwidth}
		\begin{tabular*}{\textwidth}{@{\extracolsep{\fill}}llcccccccll@{\extracolsep{\fill}}}
    \toprule
    Dataset name  & Samples & W & G & R & D & L & O & S & Data origin & Note                                                    \\     
    \midrule 
    
    \textdagger CVC-FP~\cite{Heras15} & 122 & \cmark & \xmark & \xmark & \cmark & \xmark & \cmark & \xmark & Europe region & \\ 
    \textdagger Rent3D~\cite{liu2015rent3d} & 215 & \cmark & \xmark & \xmark & \xmark & \xmark & \xmark & \xmark & Central London & \multirow{2}{*}{\parbox{2.9cm}{Include unit photos 2-30 photos / unit}} \\ \\
    R-FP~\cite{dodge2017parsing} & 500 & \cmark & \xmark & \xmark & \xmark & \xmark & \xmark & \xmark &  & Highly variable data \\ 
    *R2V~\cite{liu2017raster} & 100k+ & \cmark & \xmark & \xmark & \cmark & \xmark & \xmark & \xmark & Japan region & Original data~\cite{LIFULL15}\\ 
    \textdagger CubiCasa5K~\cite{kannala2019cubicasa5k} & 5000 & \cmark & \xmark & \cmark & \cmark & \xmark & \cmark & \cmark & Finland region & \\     
    Versailles FP~\cite{Swaileh21} & 500 & \cmark & \xmark & \xmark & \xmark & \xmark & \xmark & \xmark & Versailles Palace & \\ 
    Floor Plan CAD~\cite{Fan_2021_ICCV} & 15663 & \cmark & \xmark & \xmark & \cmark & \cmark & \cmark & \cmark &  &  \\ 
    MLSTRUCT-FP~\cite{Pizzaro2023} & 954 & \cmark & \xmark & \xmark & \xmark & \xmark & \xmark & \xmark & Chilean region & \multirow{2}{*}{\parbox{2.9cm}{Multi-unit floor plan dataset}} \\ \\
    
    \bottomrule 
    \end{tabular*}
    \end{minipage}
    \end{center}
    \caption{The datasets potentially suitable for the discussed framework; (\textdagger) evaluated in this research, * NOT publically accessible. The abbreviations stand for: W (Wall), G (Glass wall), R (Railing), D (Door), L (sLiding door), O (windOw) and S (Stairs).}
    \label{t:FPDatasets}
\end{table*}

The proposed annotations listed in~\Cref{t:FPDatasets} are based on the annotated classes outlined in \Cref{fig:qualitative_3d_multi}. If a class is not annotated as a separate one but is included in another class (for instance, a \textit{sliding door} is annotated as a \textit{door}), it is considered not annotated. The seven selected classes (\textit{wall}, \textit{glass wall}, \textit{railing}, \textit{door}, \textit{sliding door}, \textit{window} and \textit{stairs}) follow from the general requirements of the rescue services on the model representation. Most of the datasets do not annotate all of these classes, as they were designed for substantially different applications.

The datasets employed in evaluating this study (the Rent3D or R3D and the CubiCasa) were not used in the originally published form. The R3D~\cite{liu2015rent3d} was utilized in its extended variant as proposed by Zeng et al. ~\cite{zeng2019deep}. Instead of the CubiCasa5K ~\cite{kannala2019cubicasa5k}, we applied a refined version, named simply the CubiCasa; this contained fewer floor plans compared to the original, more emphasis being placed on consistency and label accuracy. Additional information may be found in ~\Cref{t:FPDatasetsUsed}.

\begin{table*}[t!]
	\begin{center}
        \begin{minipage}{\textwidth}
		\begin{tabular*}{\textwidth}{@{\extracolsep{\fill}}llll@{\extracolsep{\fill}}}
    \toprule

    Dataset name  & Samples & Resolution [px] & Augmented samples\\     
    \midrule 

    CVC-FP & 122 & w: 905-7383; h: 1027-5671 & 6966 \\ 
    Rent3D (extended) ~\cite{zeng2019deep} & 225 & w: 337-1104;
h: 175-1024 & 957 \\
    CubiCasa & 560 & w: 254-5052;
h: 206-6768 & 7265 \\ 

    \bottomrule 
    \end{tabular*}
    \end{minipage}
    \end{center}
    \caption{Evaluated dataset details, numbers of samples before and after augmentation and range of sample resolutions}
    \label{t:FPDatasetsUsed}
\end{table*}

Both of the datasets were augmented by combining flips, rotations, crops, and scaling. The data splits were set to 60, 20, and 20\% (train, validation, test). The segmentation model was also evaluated on the MURF, our multi-unit floor plan dataset, based on images of large buildings. Due to their sizes and noise levels, the floor plans in the MURF are of a high complexity compared to the CubiCasa. The MURF considers all seven semantic classes: walls, glass walls, railings, doors, sliding doors, windows, and stairs. Regrettably, this dataset could not be published herein due to regulatory compliance. Still, we decided to present the results, as the MURF dataset is highly complex and valuable. The results are exposed in \Cref{fig:qualitative_3d_multi}. More information about the used datasets is provided in the supplementary materials.

Some of the larger datasets exposed in \Cref{t:FPDatasets} were not considered in the evaluation for specific technical or legal reasons. This fact applies to the R2V dataset, which is difficult to obtain, and the Floor Plan CAD dataset, which provides the floor plans in a tiled form, with no apparent way of assembling the tiles into more complex structures that are desired for the evaluation.

\subsection{Evaluation Metrics}

To evaluate the experiments, we decided to use the recall, F1 score~\cite{chinchor1993muc}, and intersection over union (IoU)~\cite{jaccard1912distribution} metrics.

The F1 score is defined using the basic confusion matrix indicators, the correct classification TP (true positive) and TN (true negative), and the incorrect classification FP (false positive) and FN (false negative). Whereas precision penalizes the FP results and recall does so in the FN cases, the F1 score elegantly combines both, balancing them. The F1 score is high only if both the precision and the recall are also high.

The IoU is a simple metric, often used where the object shape and localization are sought (segmentation, object detection, and other aspects). Thus, we are interested in the ratio between the overlapping area of the two objects (the intersection) and the area of their union. In this study, the objects that are being analyzed are the patches of a bitmap; the intersections are then the shared pixels, and the union would be the totality of all the pixels in either object.

\subsection{Experimental Setup}

All of the investigated models were trained for a maximum of 200 epochs with a batch size of one or two due to the GPU memory limitations and the large scale of the training images. Further, a bigger batch size would not offer any benefits in terms of the model accuracy, as the group normalization performs well on small batches~\cite{wu2018group}. For the proposed architectures \modelOne{} and \modelTwo{}, the ADAM optimizer was used, the learning rate being $1\times 10^{-4}$. The experimental setup is further discussed in the supplementary materials.

The learning rate was reduced by a half after 10 epochs until we reached a minimum of $1\times 10^{-5}$ if the validation loss had been not reduced. An early stopping criterion with a patience of 30 epochs was employed as well. All of the models were pre-trained on the ImageNet dataset~\cite{deng2009imagenet}. The investigated implementations are based on the TensorFlow and were trained on a machine with $2\times$ NVIDIA RTX A5000 GPU with a 24 GB VRAM, $2\times$ dual Intel Xeon Silver 4208 CPU @ 2.10GHz 8-core processors, and 1 TB SSD storage.

\section{Experimental Results}

Quantitative result analysis was performed using the aforementioned metrics as a statistic on the pixel level. Qualitative results in terms of the individual floor plan segmentations and reconstructions were analysed to determine potential limitations.

\begin{figure*}[!t]
    \centering
    \subfloat[Input]{%
        \includegraphics[width=0.4\columnwidth]{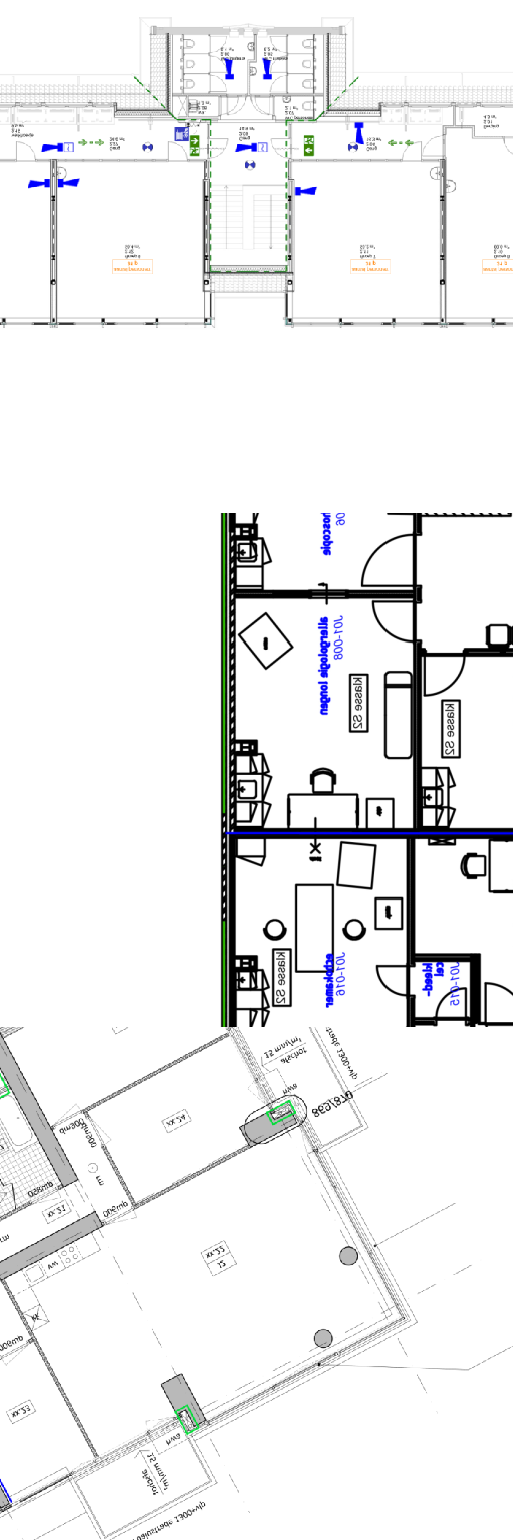}%
    }
    \hfill
    \subfloat[\modelTwo{} \label{fig:qualitative_3d_multi_rec}]{%
        \includegraphics[width=0.4\columnwidth]{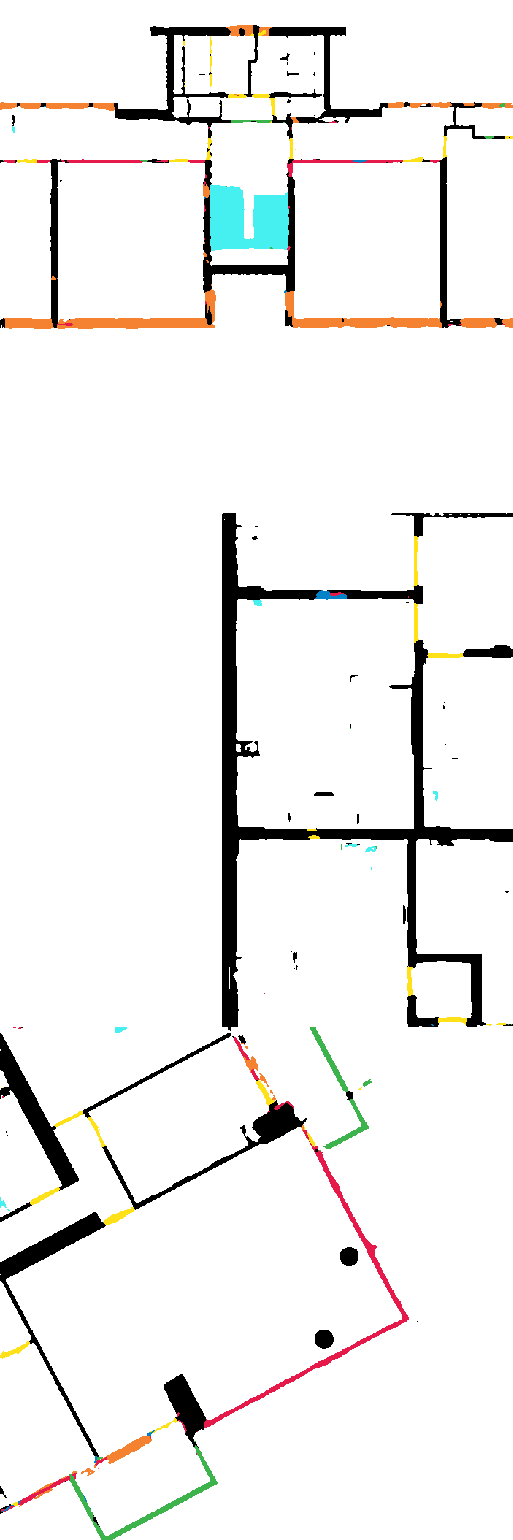}%
    }
    \hfill
    \subfloat[\modelTwo{} pp]{%
        \includegraphics[width=0.4\columnwidth]{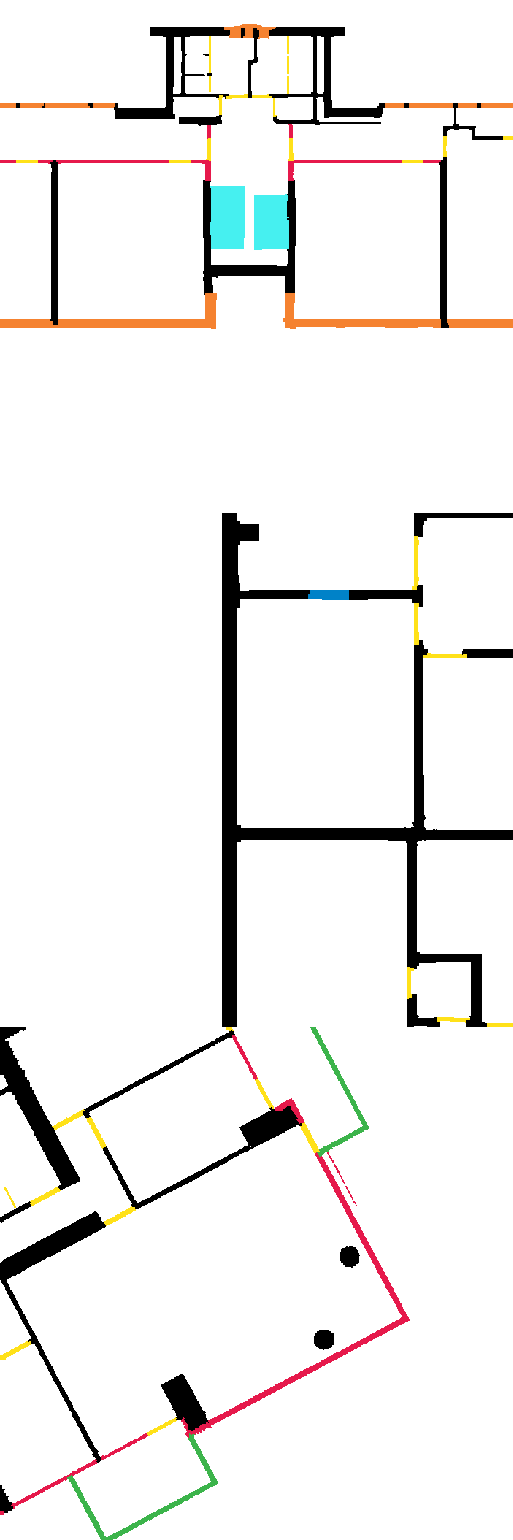}%
    }
    \hfill
    \subfloat[3D model \label{fig:qualitative_3d_multi_model}]{%
        \includegraphics[width=0.4\columnwidth]{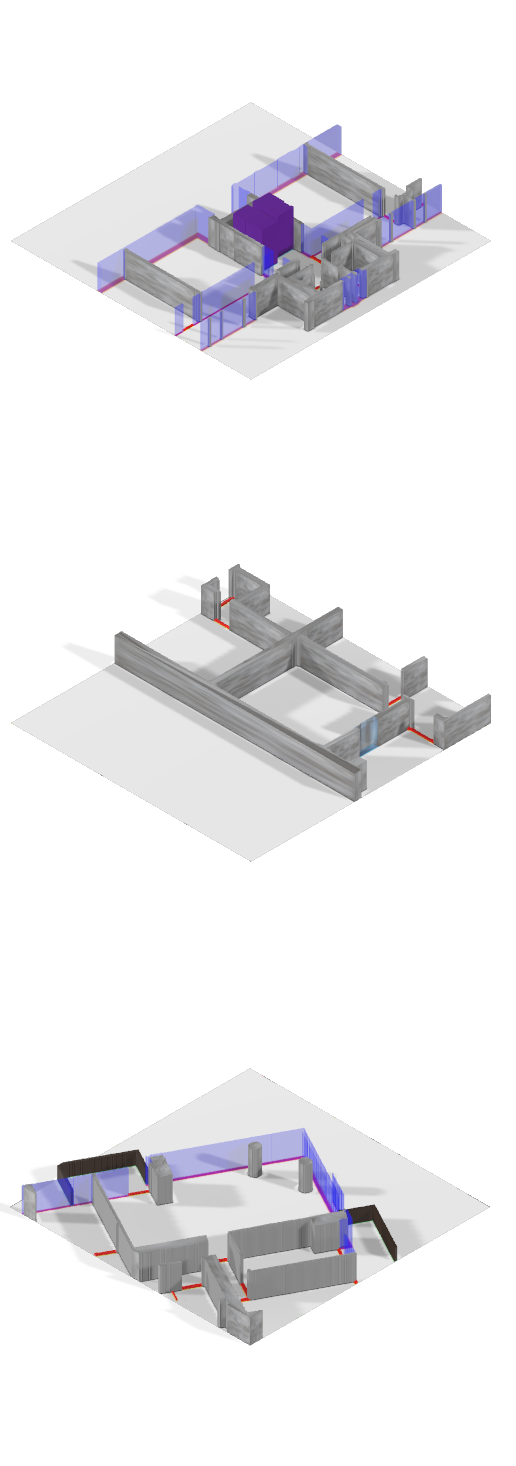}%
    }
    
    \vspace{10pt}
    \includegraphics[width=0.8\textwidth]{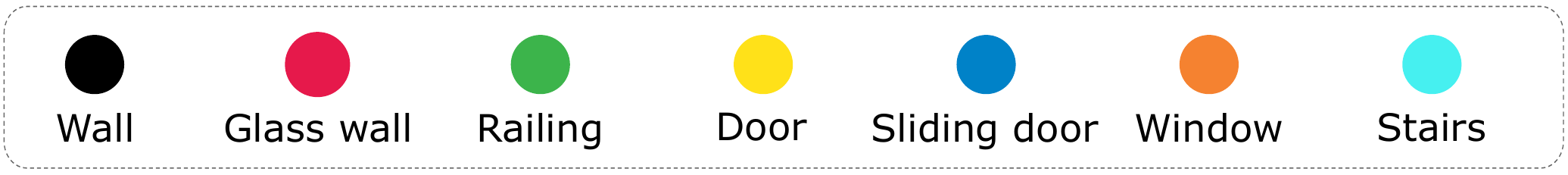}
    
    \caption{Example of several floor plans from the MURF dataset processed by the proposed framework using the \modelTwo{} segmentation technique. (a) Input samples, (b) Outputs of the recognition module, (c) Outputs of the post-processing module, (d) 3D reconstructed models.}
    \label{fig:qualitative_3d_multi}
\end{figure*}

An example of the overall results of our proposed pipeline including the segmentation, post-processing, and 3D reconstruction on the MURF dataset is shown in~\Cref{fig:qualitative_3d_multi}. More details about the achieved results over the other investigated datasets are provided in the supplementary materials.


\subsection{Comparing the Proposed Framework with the SOTA Methods}

Importantly, \Cref{t:ResMethods} compares the performance of our pipeline carrying the \modelOne{} and \modelTwo{} segmentation cores with the DFPR$_\beta$ and the Cubicasa5k$_\beta$. The proposed models performed almost the same, as the observed metrics do not differ at all. The two other models showed worse results over all of the observed classes and in all of the investigated metrics. The proposed models reached an average F1 score of 0.86 and an average IoU of 0.76 compared to the F1 score of 0.74 and IoU of 0.61 in the DFPR$_\beta$ and the F1 score of 0.74 and IoU of 0.60 in the Cubicasa5k$_\beta$.

\begin{table*}[ht!]
	\begin{center}
    \begin{minipage}{\textwidth}
		\begin{tabular*}{\textwidth}{@{\extracolsep{\fill}}lcccccccccccc@{\extracolsep{\fill}}}
        \toprule
        \multicolumn{1}{c}{} & \multicolumn{3}{c}{\modelOne{}} & \multicolumn{3}{c}{\modelTwo{}} & \multicolumn{3}{c}{DFPR$_\beta$} & \multicolumn{3}{l}{CubiCasa5k$_\beta$} \\
        & Rec.    & F1     & IoU    & Rec.   & F1     & IoU    & Rec.   & F1     & IoU    & Rec.     & F1      & IoU     \\
        \midrule
        Walls                & 0.90    & 0.90   & 0.83   & 0.90   & 0.90   & 0.83   & 0.75   & 0.83   & 0.72   & 0.80     & 0.86    & 0.75    \\
        Railings             & 0.74    & 0.77   & 0.63   & 0.76   & 0.77   & 0.63   & 0.42   & 0.53   & 0.36   & 0.42     & 0.55    & 0.38    \\
        Doors                & 0.82    & 0.82   & 0.70   & 0.82   & 0.82   & 0.70   & 0.57   & 0.69   & 0.53   & 0.57     & 0.68    & 0.52    \\
        Windows              & 0.91    & 0.90   & 0.82   & 0.92   & 0.90   & 0.82   & 0.72   & 0.81   & 0.68   & 0.75     & 0.81    & 0.69    \\
        Stairs               & 0.87    & 0.89   & 0.81   & 0.86   & 0.89   & 0.81   & 0.79   & 0.85   & 0.74   & 0.70     & 0.79    & 0.67    \\
        \hline
        \textbf{Mean}        & \textbf{0.85}    & \textbf{0.86}   & \textbf{0.76}   & \textbf{0.85}   & \textbf{0.86}   & \textbf{0.76}   & 0.65   & 0.74   & 0.61   & 0.65     & 0.74    & 0.60   \\
        \bottomrule 
		\end{tabular*}
    \end{minipage}
	\end{center}
    \caption{The pixel-wise recall, F1-score, and IoU reached with various methods over the Cubicasa dataset.}
    \label{t:ResMethods}
\end{table*}

\subsection{Evaluating the Proposed Framework on Various Datasets}

The data in \Cref{t:ResMethods} represent the results achieved in the proposed pipeline carrying the \modelOne{} and \modelTwo{} segmentation cores over various datasets. These results show a comparable and consistent performance of the proposed segmentation cores over various datasets with slightly better results achieved with the \modelTwo{} over the CVC-FP dataset.

\begin{table*}[ht!]
	\begin{center}
        \begin{minipage}{\textwidth}
            \begin{tabular*}{\textwidth}{@{\extracolsep{\fill}}lcccccccccc@{\extracolsep{\fill}}}
            \toprule
            \multicolumn{2}{c}{} & \multicolumn{3}{c}{Cubicasa} & \multicolumn{3}{c}{R3D} & \multicolumn{3}{c}{CVC-FP} \\
            & & Rec.     & F1   & IoU  & Rec. & F1   & IoU  & Rec.   & F1   & IoU  \\
            \midrule
            \multirow{2}{*}{\modelOne{}} & Walls    & 0.90 & 0.90 & 0.83    & 0.94 & 0.94 & 0.90   & 0.95 & 0.94 & 0.88 \\
                                 & Mean    & 0.85 & 0.86 & 0.76    & 0.89 & 0.90 & 0.82   & 0.87 & 0.86 & 0.77 \\
            \midrule
            \multirow{2}{*}{\modelTwo{}} & Walls    & 0.90 & 0.90 & 0.83    & 0.94 & 0.94 & 0.90   & 0.96 & 0.95 & 0.91 \\
                                 & Mean    & 0.85 & 0.86 & 0.76    & 0.89 & 0.90 & 0.82   & 0.91 & 0.90 & 0.82 \\
            \bottomrule 
		\end{tabular*}
        \end{minipage}
	\end{center}
    \caption{Results of the proposed pipeline with \modelOne{} and \modelTwo{} segmentation models reached over Cubicasa, R3D, and CVC-FP datasets. As the R3D datasets do not contain all classes evaluated in~\Cref{t:ResMethods}, only walls and mean results are compared.}
    \label{t:ResMDatasets}
\end{table*}

\section{Conclusions}
\label{sec:discussion}

This paper was conceived to seek a method for generating a digital twin (in the form of a 3D model) to a real-world building from less-than-ideal floor plans. Having a reliable 3D representation of a building rather than collections of 2D plans in various forms and quality is an important step towards safer and more efficient emergency services in the ever-growing urban environment.

We developed a pipeline that not only analyzes an input drawing at the pixel level but also segments and classifies selected features to generate a 3D reconstruction of the relevant building. Two new multi-scale, fully trainable pixel segmentation models designed for multi-unit floor plan analysis, the \modelOne{} and \modelTwo{}, are proposed as well. 
The pipeline outperformed the SOTA DFPR and Cubicasa5k methods, and it was were evaluated on three datasets, reaching consistently high evaluation metrics over various data.

By design, the recognition part lacks a region of interest (ROI) detection module, which otherwise allows cropping the floor plan; such structuring generally improves the accuracy. The main reason for not including the module rests in that object detection models, such as those in the YOLO family, are sufficient enough to crop floor plans to their appropriate size, as shown in, for example,~\cite{lv2021residential}, where the authors employ a variant of the YOLOv4 for their ROI module.

A comparison of the pipeline with two SOTA floor plan processing models, the DFPR and CubiCasa5K, showed that our approach surpasses these techniques on the CubiCasa dataset. We assume that the increased complexity of our pipeline allows and will enable us to capture more effectively a larger range of features necessary to identify the elements in floor plans of different sizes.

More recent SOTA floor plan processing frameworks presented in~\cite{lv2021residential}, \cite{electronics10222729}, and~\cite{9893304} unfortunately do not have their source codes publicly available, so the comparison in the same manner as with DFPR and CubiCasa5k was not possible. Nevertheless, the authors of~\cite{9893304} tested their framework on the R3D dataset, where they reached a recall of 0.89 in the case of walls. Our approach performed better in this case with a resulting recall of 0.94 in the case of wall recognition. Results for other methods and experiments were not comparable to those presented in this paper (different datasets, different metrics). 

A limitation of the proposed reconstructing method lies in the visualization of stairs. Currently, stairs are visualized via polygons of an equal height, set to a higher value than the other semantic classes. Such detection, however, does not consider the orientation of the stairs, which would further improve the generated 3D models. Another constraint rests in visualizing round-shaped objects. At present, the segmentation masks are transformed into polygons, the endpoints of each connected component functioning as the vertices. Disadvantageously, this approach generates stair-like artifacts due to the discrete values in the segmentation mask, and the problem is especially apparent in round-shaped objects. Besides a further analysis and tackling of these challenges, our future work will also focus on the processing of multi-floor plans, this will require implementation of methods to align the individual storey floor plans, which will allow understanding the context of stairwells and lift shafts across the building.

As the presently available datasets usually contain the floor plans of smaller units such as flats, we will further focus our future research on collecting a dataset consisting of large floor plans across various types of buildings, including, public buildings, and medical facilities. We also intend to improve our segmentation network to process older plans, which are available only on paper sheets drawn according to various technical standards. This improvement is assumed to markedly refine the development of the DT, as the older buildings can be included in such models.

In the supplementary materials is provided also an ablation study evaluating the contribution of all parts of the proposed pipeline.


\bibliographystyle{IEEEtran} 
\bibliography{references}

\begin{IEEEbiography}[{\includegraphics[width=1in,height=1.25in,clip,keepaspectratio]{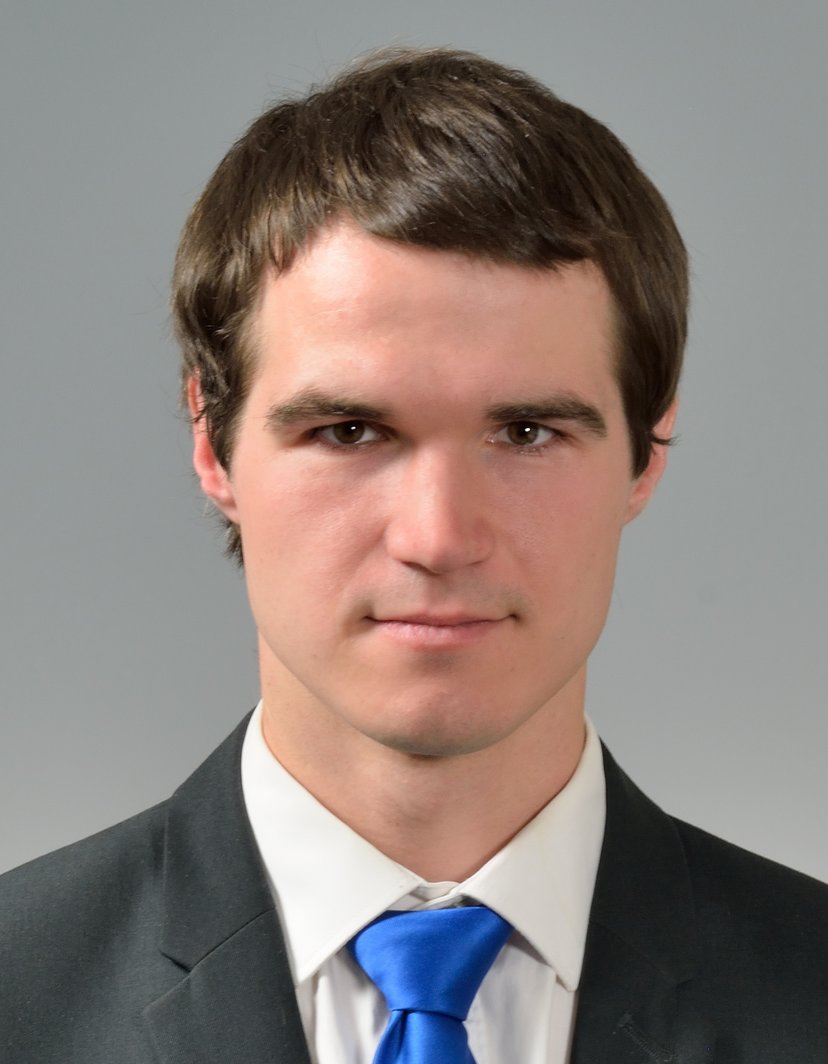}}]{Lukas Kratochvila} is a PhD student in the Machine Vision Group of the Department of Control and Instrumentation, FEEC, Brno University of Technology. He graduated with his Master's within the same group in 2019. The aim of his research is visual object detection, motion tracking and the hardware for the AI tasks. He took part in the Erasmus + program and gained countless experiences at the Norwegian University of Science and Technology in Trondheim; for example, in the image style transfer using the CNNs.
\end{IEEEbiography}

\begin{IEEEbiography}[{\includegraphics[width=1in,height=1.25in,clip,keepaspectratio]{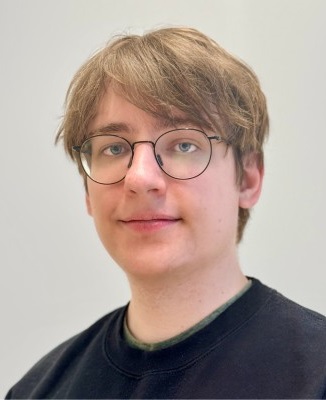}}]{Gijs de Jong} is a dedicated software engineer with several years of experience building scalable game servers. Currently, he is interested in language design and microservice architectures.
\end{IEEEbiography}

\newpage


\begin{IEEEbiography}[{\includegraphics[width=1in,height=1.25in,clip,keepaspectratio]{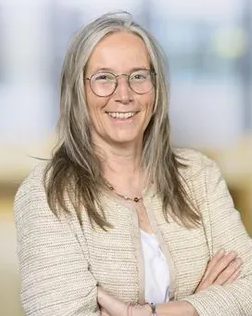}}]{Monique Arkesteijn}
is specialised in Corporate Real Estate (CRE) alignment. CRE is different from commercial real estate, which focuses on return on investment (supply side); whereas CRE supports the primary process (demand side). A long-standing issue in CRE is the alignment of an organization’s real estate to its corporate strategy, where many opt for optimum added value. In CRE decision making, traditionally there is a strong financial focus, thus making other non-financial value indefensible. In her work she is using the stakeholder approach, taking into account all values both financial, objective and subjective, intangible values, and combining them in a new single-valued objective function. The innovative Preference-based Accommodation Strategy approach (PAS) that she developed in her PhD made value defensible by redefining value as technically equivalent to preference and using a design and decision approach for the alignment problem.  PAS is a major improvement to CRE alignment because stakeholders’ values are integrated explicitly in the algorithm and separate from the objective measurements (as CO2 emission or m2). and thereby creating the best (working) environment for people. 
\end{IEEEbiography}

\begin{IEEEbiography}[{\includegraphics[width=1in,height=1.25in,clip,keepaspectratio]{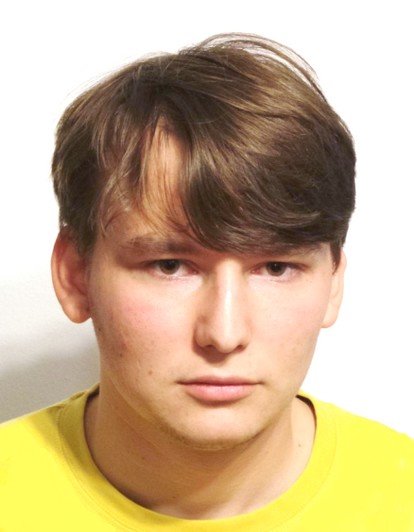}}]{Tomas Zemcik}
is a PhD student in the Machine Vision Group of the Department of Control and Instrumentation, FEEC, Brno University of Technology. He graduated with his Master's within the same group and department in 2019. For five years he has been with CAMEA, spol. s r.o. (a company specialising in image and signal processing in industry and traffic applications) as a developer. His experience also includes involvement in several projects including V3C. During his graduate studies he participated in the Erasmus+ programme and spent a year at the Tampere University of Technology. 
\end{IEEEbiography}

\begin{IEEEbiography}[{\includegraphics[width=1in,height=1.25in,clip,keepaspectratio]{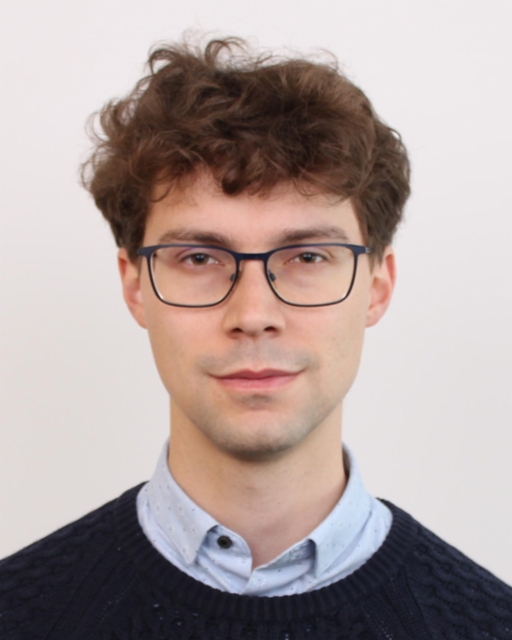}}]{Simon Bilik}
received the PhD degree in technical cybernetics and computer vision from the Department of Control and Instrumentation, Brno University of Technology, Czech republic and the Computer Vision and Pattern Recognition Laboratory, LUT University, Finland, in 2024. His research field includes applied machine vision and machine learning. Currently, he works as researcher at the Institute for Research and Applications of Fuzzy Modeling of the University of Ostrava together with the research position at Department of Informatics of the Mendel University in Brno.
\end{IEEEbiography}

\begin{IEEEbiography}[{\includegraphics[width=1in,height=1.25in,clip,keepaspectratio]{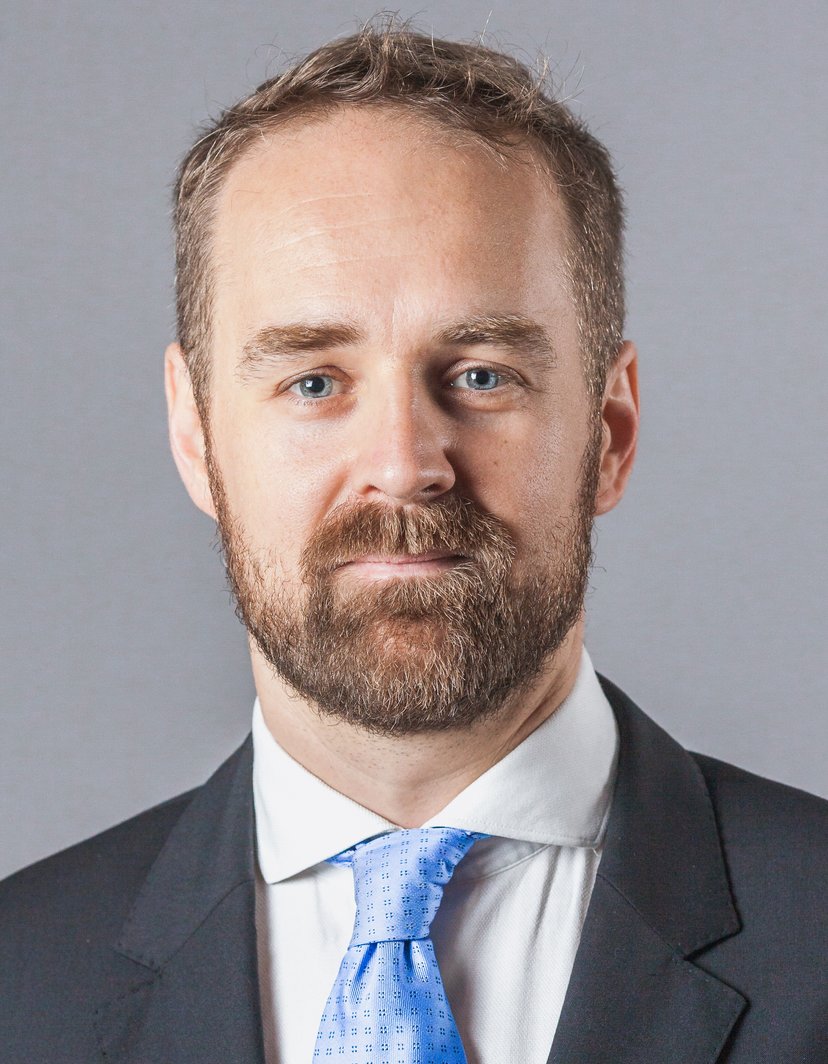}}]{Karel Horak}
is Ass. Prof. in the Department of Control and Instrumentation at Brno University of Technology. His expertise is of computer vision, optics, machine learning and signal processing. He obtained his master and PhD degrees in cybernetics in 2004 and 2008 respectively and he is the head of Machine Vision Group since 2011. He obtained a Top10 excellence award at BUT and Czech Engineering Academy award for example.  He has been as a research fellowship at Technischen Universität Wien and at Austrian Institute of Technology in Austria and at Lappeenranta-Lahti University of Technology in Finland.
\end{IEEEbiography}

\begin{IEEEbiography}[{\includegraphics[width=1in,height=1.25in,clip,keepaspectratio]{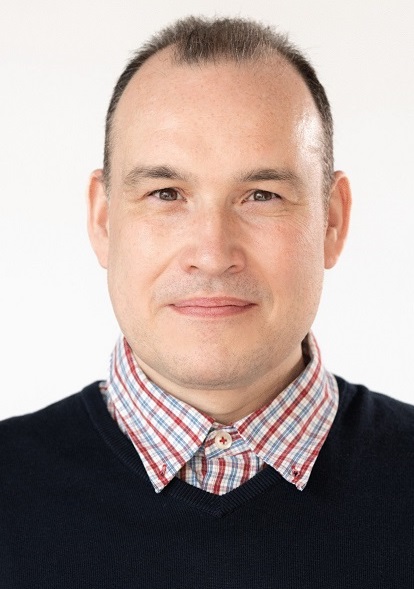}}]{Jan S. Rellermeyer}
received his MSc CS in Distributed Systems from ETH Zürich, Switzerland, in 2006 and completed his PhD in Computer Science in the Systems Group at ETH in 2011. His thesis work was on Modularity as a Systems Design Principle. In 2010, he did an internship at Microsoft Research in Redmond, WA in the eXtreme Computing Group and worked on an operating system design and energy management for the Intel SCC (Rock Creek) architecture. After graduation, he joined IBM Research in Austin, TX where he worked as an RSM in the Future Systems and Next Generation Datacenters Group. He was part of the team that built and released the first mobile app that IBM ever published, IBM Mobile Systems Remote. This effort lead sparked the IBM Mobile First initiative. He was also a co-lead of the winning 2013 Global Technology Outlook (GTO) topic Software-Defined Environments. He then worked on workload-optimized systems for the POWER architecture where he leveraged features like coherently-attached flash through CAPI. This work is available on Github, In 2016, he worked on cloud and datacenter computing and in 2017 his group switched to working on machine learning and natural language processing. In 2013, he became an Adjunct Assistant Professor in the Department of Computer Science at The University of Texas at Austin and taught the Programming Languages, Programming Languages Honors, and Principles of Computer Systems classes until 2017. In September 2017, He left IBM Research and joined the Distributed Systems Group at TU Delft as an Assistant Professor. He was the Invited Researcher of the OSGi Alliance from 2008-2011 and contributed to several successful open source projects in the Eclipse and Apache Foundations. He was the project lead of the Eclipse Concierge project. In 2015, he was a mentor for the Facebook Open Academy and worked with students from UNICAMP Brazil. In 2017, my co-authors Gustavo Alonso, Timothy Roscoe, and he received the ACM/IFIP/USENIX Middleware Test of Time award for our 2007 paper R-OSGi: Distributed Applications through Software Modularization. In 2019, he was appointed as a Master Coordinator for the Software Technology Track of the MSc in Computer Science program at TU Delft. 
\end{IEEEbiography}

\EOD

\end{document}